\documentclass[11pt]{article}
\usepackage{acl2016}
\usepackage{times}
\usepackage{url}
\usepackage{graphicx}
\usepackage{latexsym}
\usepackage{enumitem}% http://ctan.org/pkg/enumitem
\usepackage{algorithm}
\usepackage{algorithmic}
\usepackage{color}

\aclfinalcopy % Uncomment this line for the final submission
\title{DialPort: Connecting the Spoken Dialog Research Community to Real User Data}

\author{Tiancheng Zhao\thanks{\indent Both authors equally contributed to this work} $^{1}$ Kyusong Lee$^{*1,2}$ Maxine Eskenazi$^1$ \\\\
  $^1$Language Technologies Institute, Carnegie Mellon University \\
  $^2$Pohang University of Science and Technology\\
  $^1${\tt \{tianchez,max+\}@cs.cmu.edu}, $^2${\tt kyusonglee@postech.ac.kr}}

\begin{document}
\maketitle
\begin{abstract}
  This paper describes a new spoken dialog portal that connects systems produced by the spoken dialog academic research community and gives them access to real users. We introduce a distributed, multi-modal, multi-agent prototype dialog framework that affords easy integration with various remote resources, ranging from end-to-end dialog systems to external knowledge APIs. To date, the DialPort portal has successfully connected to the multi-domain spoken dialog system at Cambridge University, the NOAA (National Oceanic and Atmospheric Administration) weather API and the Yelp API.
\end{abstract}

\section{Introduction}
The advent of Siri, Cortana and other agents has generated interest in spoken dialog research. These applications have sparked the imagination of many and led them to believe that speaking to intelligent agents is useful. The research community needs to profit from this interest by creating a service for the general public that can gather real user data that can be used to make dialog systems more robust. It can also be used to carry out experiments on comparative studies. Industry already has access to large data sets and sometimes to pools of real users that are viewed as strategic competitive resources and so not shared with the research community. Yet much fundamental research remains to be done, such as signal processing in noisy conditions, recognition of groups of difficult users (like the elderly and non-natives), management of complex dialogs (such as multi party meetings, negotiations, and multimodal interaction), and the automatic use of meta linguistic information such as prosody. The academic community needs to unite through one common portal to be able to gather enough real user data to make significant impact. It is extremely difficult for any one group to devote time to collecting a significant amount of real user data. The users must be found and kept interested and the interface must be created and maintained. One data gathering portal that all dialog systems can be connected to gives potential users a variety of interesting applications, much in the way that virtual assistants do not only provide information about scheduling. The DialPort portal was created for this purpose. \\
\indent The Dialog Research Center at Carnegie Mellon (DialRC) has given the community real user data from the Lets Go System~\cite{raux2005let} as well as access to the system to run studies. But research is carried out in other areas, beyond  simple form filling. Just as one research group cannot attract a diverse pool of regular users, one group cannot cover all of the possible applications, such as virtual humans and robots with multimodal communication. Thus the goal of DialPort is to attract and maintain a pool of real users to a group of spoken dialog applications. \\
\indent The first year goal is to create the portal and link it to other systems. Once the working portal can give a variety of useful information, a service such as Prefinery\footnote{https://www.prefinery.com/} will be used to attract the real users. These services solicit potential users, giving bonuses for signup and usage as well as for getting friends to sign up. In this paper, we present the DialPort portal that will link many different research applications and will provide real user data. Section~\ref{sec:prior} discusses related work; Section~\ref{sec:architecture} describes the overall architecture; Section~\ref{sec:agent},~\ref{sec:reinforest} and~\ref{sec:chatbot} discuss the core modules; Section~\ref{sec:protocol} explains the integration protocol; Section~\ref{sec:current} reviews the current progress and Section~\ref{sec:conclude} concludes.

\section{Related Work}
\label{sec:prior}
%There are two main branches of research in Spoken dialog systems (SDS): goal-driven dialog systems and non-goal-driven dialog systems. The first category includes dialog domains that have specific goals for the agent to achieve, such as providing bus schedule information, recommending restaurants or giving weather updates~\cite{raux2005let,bobrow1977gus}. Many goal-driven dialog management frameworks obtain dialog polices that can achieve the their goals via natural conversation with users. Popular approaches include framed-based methods~\cite{glass1999real}, agenda-based methods~\cite{bohus2009ravenclaw} and Partially Observable Markov Decision Process methods~\cite{williams2007partially}. 

%Non-goal-driven dialog systems focus on the chatting behavior with human users. A non-goal-driven dialog system is usually used for entertainment or as the back-off strategy for a goal-oriented dialog system~\cite{wallace2009anatomy,weizenbaum1966eliza,banchs2012iris}. DialPort combines both approaches by using non-goal driven dialog as a back-off strategy for a goal-oriented agent. The following two scenarios illustrate the usefulness of a non-goal-driven dialog system. Scenario 1: the users' utterances are out-of-domain and the dialog agent needs to recover from non-understanding errors. Secnario 2: the users ask questions that require a open-domain knowledge base (e.g. search engine)  such as "who founded Microsoft?" or "How much is an iPhone 6s?" Section~\ref{sec:chatbot} deals with the constructions of non-goal driven dialog systems.
Several SDS-building industrial platforms have recently become available to non-expert developers. Microsoft Research released Language Understanding Intelligent
Service (LUIS)~\cite{williams2015fast} which helps software developers create cloud-based, machine-learning powered, language understanding models for specific application domains. Facebook is building an AI platform that helps developers create chat bots that can converse in natural language\footnote{https://wit.ai}. The HALEF (Help Assistant–Language-Enabled and Free) framework from ETS leverages different open-source components to form an SDS framework that is modular and industry-standard-compliant~\cite{suendermann2015halef}. Although these platforms have helped researchers build robust SDSs more efficiently, the data that they collected has not been shared with the academic research community. \\
\indent Most early SDS work focused on single-domain SDSs, such as bus schedules, restaurant, or museum information, etc. A single-domain dialog system has limited semantic coverage. Thus, multi-domain dialog systems have appeared, enabling one system to handle several domains. Past approaches usually followed a two-stage framework~\cite{komatani2009multi,nakano2011two}, in which the first stage classifies the domain and the second stage forwards the user's request to the relevant single-domain dialog manager. This method has shown promising results for scaling up dialog systems to handle multiple domains. DialPort differs from previous frameworks in this area by proposing the concept of a \textit{multi-agent dialog system}. This system combines both goal-driven and non-goal-driven dialog agents that have been independently developed by different research teams.  The task of DialPort is to judiciously assign the user's utterance to the most relevant dialog agent and to carry out complex nested conversations with real users. The long term goal is to enable any research group to connect their SDS to DialPort using a lightweight integration protocol. DialPort makes it easy for real users to access many state-of-the-art dialog system services all in one place through a universal web-based entry point.

\section{Architecture}
\label{sec:architecture}
\begin{figure*}[ht]
\centering
    \includegraphics[width=12cm]{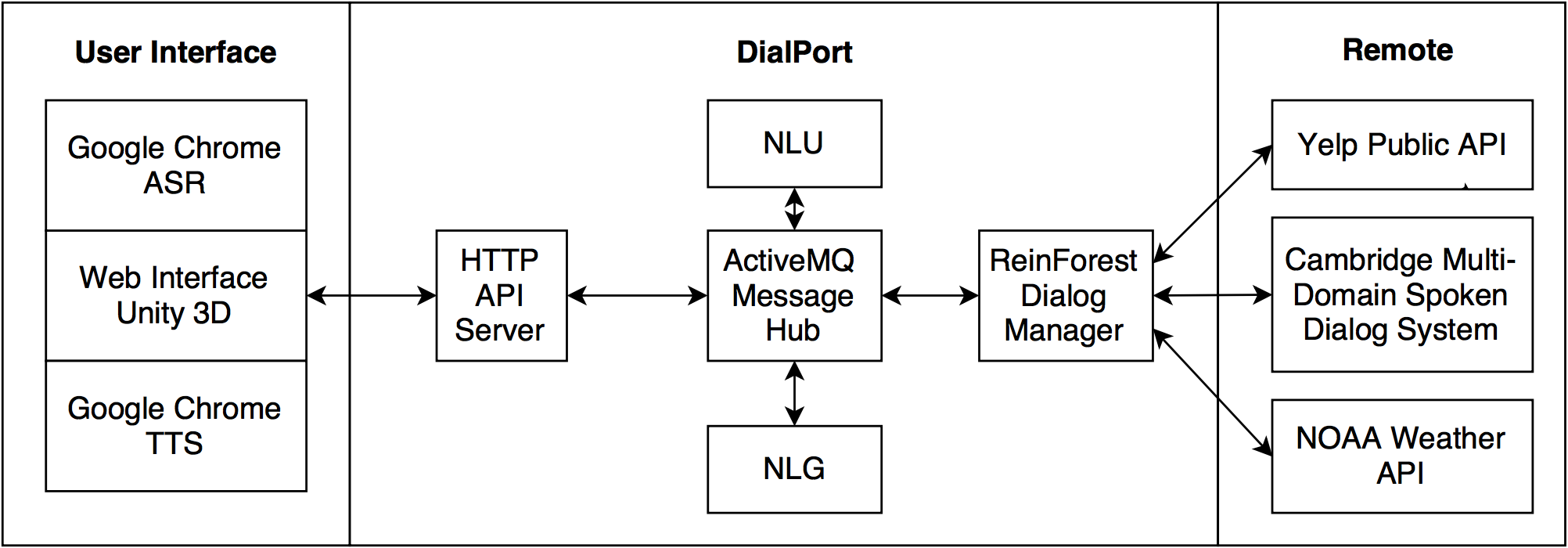}
    \caption{The overall architecture of DialPort.}
    \label{fig:sys_overall}
\end{figure*}

Figure~\ref{fig:sys_overall} presents the system architecture, comprised of three sections: User Interface, DialPort and Remote Agents.

\subsection{User Interface}
The user Interface is the publicly available front end for real users\footnote{https://skylar.speech.cs.cmu.edu}. It is in charge of both the visual and audio agents representing each dialog system. The visual representation uses WebGL Unity 3D, which will be discussed in Section~\ref{sec:agent}. The audio side uses the Google Chrome Speech ASR API to transform the user's speech into text and the Google Chrome TTS API to convert DialPort's text output into English speech.
\subsection{DialPort}
DialPort is scalable and distributed. Its central message broker is ActiveMQ, a well-known open source message broker. ActiveMQ allows us to easily connect multiple components in order to create a larger system. Building on ActiveMQ, DialPort has four main modules: the HTTP API Server, the Natural Language Understanding (NLU), the ReinForest Dialog Manager (DM) and the Natural Language Generation (NLG). With the exception of the ReinForest DM, the modules are RESTful (Representational state transfer) web services: they do not consider any state information when handling requests. All contextual information about a dialog is maintained by the ReinForest DM. The HTTP API Server is the front gate of DialPort. It converts the incoming HTTP messages into proper ActiveMQ messages and sends them to the NLU. The NLU outputs a semantic frame that contains the original utterance along with: entities, an intent and a domain. Assuming the user utterance is: "Recommend a restaurant in Pittsburgh", an example output from the NLU is given in Table~\ref{tbl:nlu}. Given the user input annotated by the NLU, ReinForest updates its internal dialog state and generates the next system response. The response is in the format of a list of dialog acts and content value tuples, $a_{sys} = [(DA_0, v_0), ... (DA_k, v_k)]$. The NLG is responsible for transforming $a_{sys}$ to the natural language surface form. Given the utterance in the previous example, an example ReinForest response is:
$a_{sys}$ \textit{= [(CONFIRM, value=Pittsburgh), (ASK, value=food\_type)]}. And the final system output is "I believe you said Pittsburgh. What kind of food do you want?"  
\begin{table}[ht!]
    \centering
    \begin{tabular}{p{0.08\textwidth}|p{0.35\textwidth}} \hline
    Domain & \{Restaurant: 0.95; Hotel: 0.05\}\\ 
    Intent & \{Request: 0.9; Inform: 0.1\} \\ 
    Entities & \{Type: Location; Value: Pittsburgh\} \\ \hline
    \end{tabular}
    \caption{Example Semantic Frame Output from the NLU.}
    \label{tbl:nlu}
\end{table}

\begin{figure*}[ht]
\centering
    \includegraphics[width=14cm]{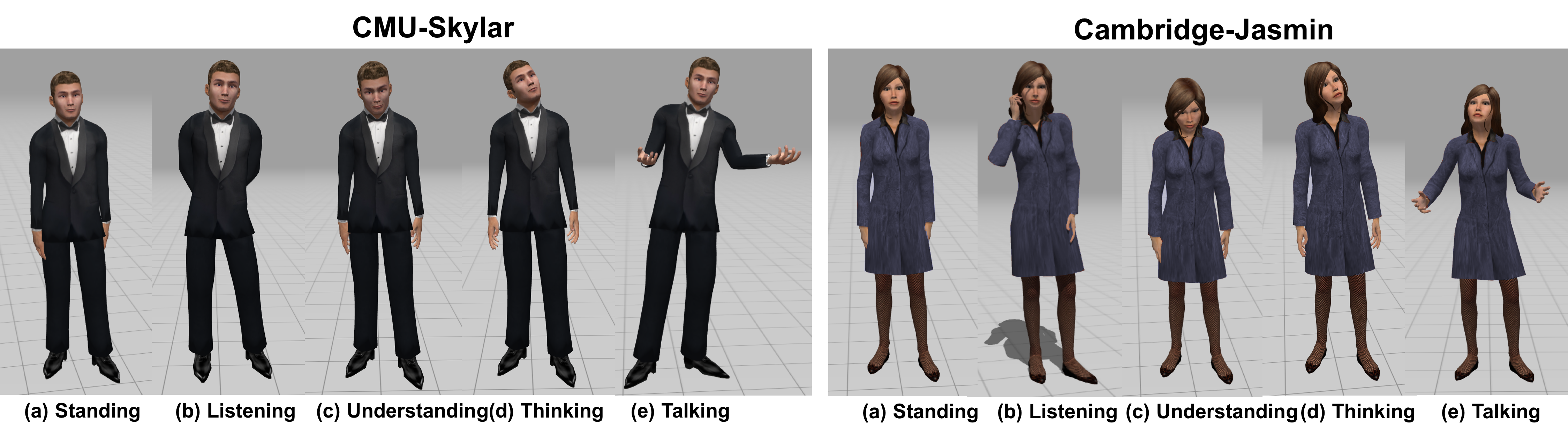}
    \caption{Non-verbal Generation of each agent - some positioning possibilities}  
    \label{fig:skylar}
\end{figure*}

\subsection{Remote Agents}
Easy integration with remote Agents is a major contribution of the proposed architecture. We define a remote agent as any external resource that can be integrated into the DialPort ecosystem. Generally there are three types of remote agents: audio remote agent, text remote agent and knowledge remote agent. 

\textbf{Audio Remote Agent}: this is a self-sustaining spoken dialog system that only has a public audio API. Therefore, an audio remote agent expects audio streaming input and returns an audio clip that contains the system's spoken response. DialPort does not presently support this type of remote agent due to the difficulty of dealing with real-time audio streaming amongst remote servers. This will be dealt with when connection to a system of this type is proposed.

\textbf{Text Remote Agent}: this type of agent provides text API, which inputs the latest ASR text output and returns the next system response in text form. It should be noted that even end-to-end spoken dialog systems can belong to text remote agents. They only need to provide a text-based API. The Cambridge Multi-domain spoken dialog system in Figure~\ref{fig:sys_overall}~\cite{gasic2015distributed} is one example. It has its own VoIP audio server and also provides a text-based API. Therefore, when the Cambridge system connects with DialPort, the latter sends the transcribed speech to Cambridge's text-based API and bypasses its VoIP server. 

\textbf{Knowledge Remote Agent}: the third type of remote agent is an external knowledge base, ranging from a web API (e.g. Yelp API) to an in-house relational database (e.g Pittsburgh bus schedule). DialPort is in charge of all of the dialog processing and uses the knowledge remote agent as the back-end.

\section{Virtual Agent}
\label{sec:agent}
This section introduces Skylar and Jasmin. Skylar is the virtual agent for DialPort. Jasmin is the Cambridge University dialog system agent.  Both agents interact with users via a web speech interface and have 3D animated embodiments powered by the Unity 3D Engine\footnote{http://unity3d.com/}. \\
\indent The way in which non-verbal expression (agent behavior) is handled is important. Users need to be able to easily interpret what the agents' current and next status are. DialPort uses a variety of character animations such as (a) standing, (b) listening, (c) understanding, (d) thinking, and (e) talking (Figure~\ref{fig:skylar}), so that users implicitly know when they should talk or should wait for the ASR results. Moreover, the non-verbal expressions of a virtual agent indicate which agent the system thinks the user is talking to. Generating natural non-verbal expression is an open research problem. At present, non-verbal expressions are manually encoded using the Mixamo engine\footnote{https://www.mixamo.com} as the following: 
\begin{itemize}\setlength\itemsep{-\parsep}    
    \item When nothing is happening, he/she is standing with his/her arms lowered to his/her sides. (Figure~\ref{fig:skylar}-a).
    \item When a user starts to talk to Skylar, he puts his hands/arms behind his back and inclines his head slightly forward and down. Jasmin puts a hand to her ear, indicating that she is listening (Figure~\ref{fig:skylar}-b).
    \item When he/she has finished listening, he/she quickly nods twice to indicate that he/she has understood and is starting to process the turn (Figure~\ref{fig:skylar}-c)
    \item After he/she nods, he/she brings his/her arms back down to his/her sides and raises his/her head, looking up at the ceiling - this indicates that the agents is thinking of a response (Figure~\ref{fig:skylar}-d).
    \item when he/she is speaking, he/she lowers his/her head and his/her gaze to look at the user and uses various hand gestures.  (Figure~\ref{fig:skylar}-e)
\end{itemize}

\section{The ReinForest Dialog Manager}
\label{sec:reinforest}
\begin{figure*}[ht]
\centering
    \includegraphics[width=15cm]{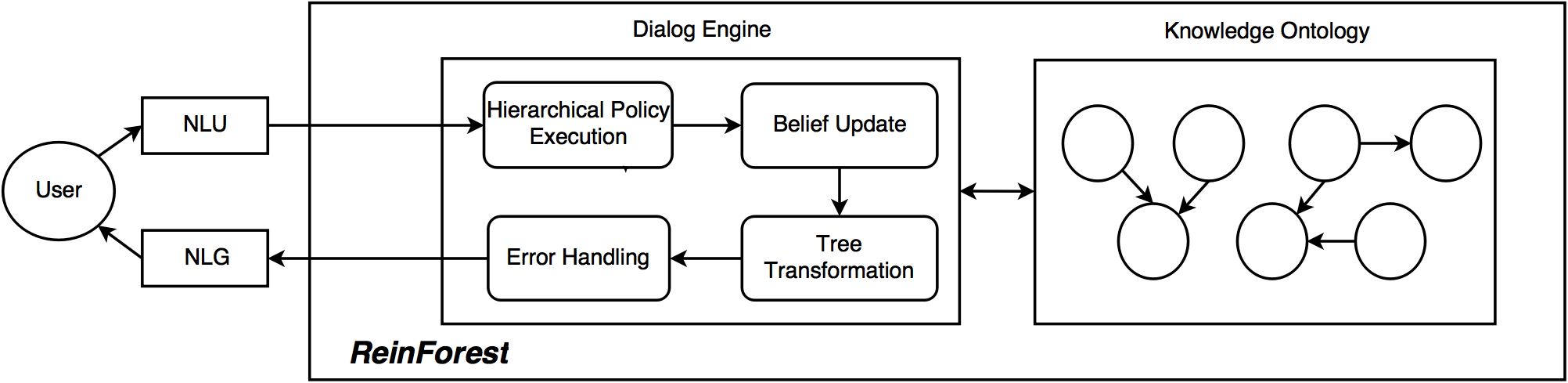}
    \caption{Overview of Reinforest Dialog Manager.}
    \label{fig:reinforest}
\end{figure*}

The challenges facing the DialPort dialog manager are that it must 1) support easy extension to a large variety of domains and 2) support mixed initiative and mixed (non)-goal driven dialogs. Specifically to deal with these two challenges, we have developed a new dialog manager based on RavenClaw~\cite{bohus2003ravenclaw}. The overview of the ReinForest DM is shown in Figure\ref{fig:reinforest}.
The core of ReinForest has two parts: the \textit{knowledge ontology} and the \textit{dialog engine}. The \textit{knowledge ontology} is a domain-dependent knowledge graph that developers can use to quickly encode domain knowledge and relations between various concepts. The \textit{dialog engine} is a domain-independent execution mechanism that generates the next system response given the current dialog state. The following sections formally define these two components.

\subsection{Knowledge Ontology}
The \textit{Knowledge Ontology} can be thought of as fuel for ReinForest. The basic unit of the ontology is a \textit{concept}, which is an abstraction of knowledge. For example, knowledge about the weather is a \textit{concept}. A \textit{concept} can have a number of dependent concepts that encode the causal relationship. In the weather domain for example, the weather concepts depend on the location and date time concepts. Therefore, in order to give weather information, the system has to have already acquired the values of date time and location. More importantly, each remote agent is also represented as a concept. For example, the Cambridge Multi-Domain Dialog System is represented as a single concept that contains information about all of the domains that it covers. 

\begin{figure}[ht]
\centering
    \includegraphics[width=0.42\textwidth]{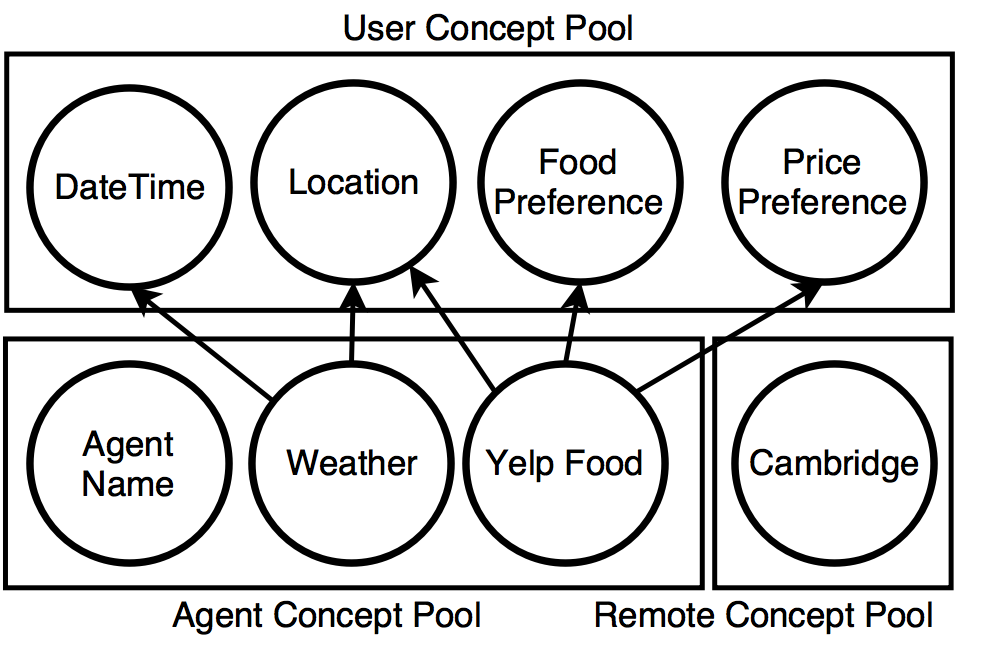}
    \caption{Example concept pools.}
    \label{fig:pool}
\end{figure}

\textbf{Concept Pool:} Given the definition of a \textit{concept}, ReinForest enables developers to construct domain knowledge in the form of a directed acyclic graph (DAG). ReinForest also introduces the idea of a \textit{concept pool} to create groups of concepts. Figure~\ref{fig:pool} gives a simple example of a \textit{concept pool} for a slot-filling dialog manager. Essentially a \textit{concept pool} is a collection of concepts that share some common properties. There are three concept pools in ReinForest: the \textit{agent concept pool}, the \textit{user concept pool} and the \textit{remote concept pool}. The \textit{agent concept pool} consists of concepts that are powered by knowledge remote agents, such as the agent's name and the weather information. The \textit{user concept pool} contains concepts that only users know, such as the date that the user is asking about, the user's names etc. Finally the \textit{remote concept pool} contains all the text remote agents, where each is considered to be a black box that can generate the next system response given the context of certain domains.

\subsection{Dialog Engine}
As illustrated in Figure~\ref{fig:reinforest}, the \textit{dialog engine} of ReinForest is a domain-independent execution mechanism that consists of 4 modules: hierarchical policy execution, belief update, tree transformation and error handling. Algorithm~\ref{alg:main} shows the pseudo code of the main execution loop inside ReinForest. The \textit{dialog engine} establishes its connection to the \textit{knowledge ontology} via the dialog state, $s$, which captures all the information about the ongoing dialogs. Next we formally define these four modules.

\begin{algorithm}
\caption{ReinForest Main Loop}
\begin{algorithmic} 
\label{alg:main}
\WHILE{$dialog.end() \neq True$}
    \WHILE{$dialog\_stack.top() \in O $}
        \STATE $execute(dialog\_stack.pop())$
    \ENDWHILE
    \IF{user has input}
        \STATE $belief\_update()$
        \STATE $tree\_transformation()$
        \STATE $error\_handle()$
    \ENDIF
\ENDWHILE
\end{algorithmic}
\end{algorithm}

\subsection{Hierarchical Policy}
Hierarchical policy has been studied extensively in the literature of both hierarchical reinforcement learning (HRL)~\cite{dietterich2000hierarchical,parr1998reinforcement,sutton1999between} and plan-based dialog management~\cite{bohus2003ravenclaw,bohus2009ravenclaw}. The contribution of ReinForest is that it formalizes plan-based dialog management in the language of HRL, which opens up the possibility of applying well-established HRL algorithms to optimize the operations of the plan-based dialog manager. We first introduce the notations of HRL and then define the dialog task tree using that language. 

\begin{figure*}[ht]
\centering
    \includegraphics[width=11cm]{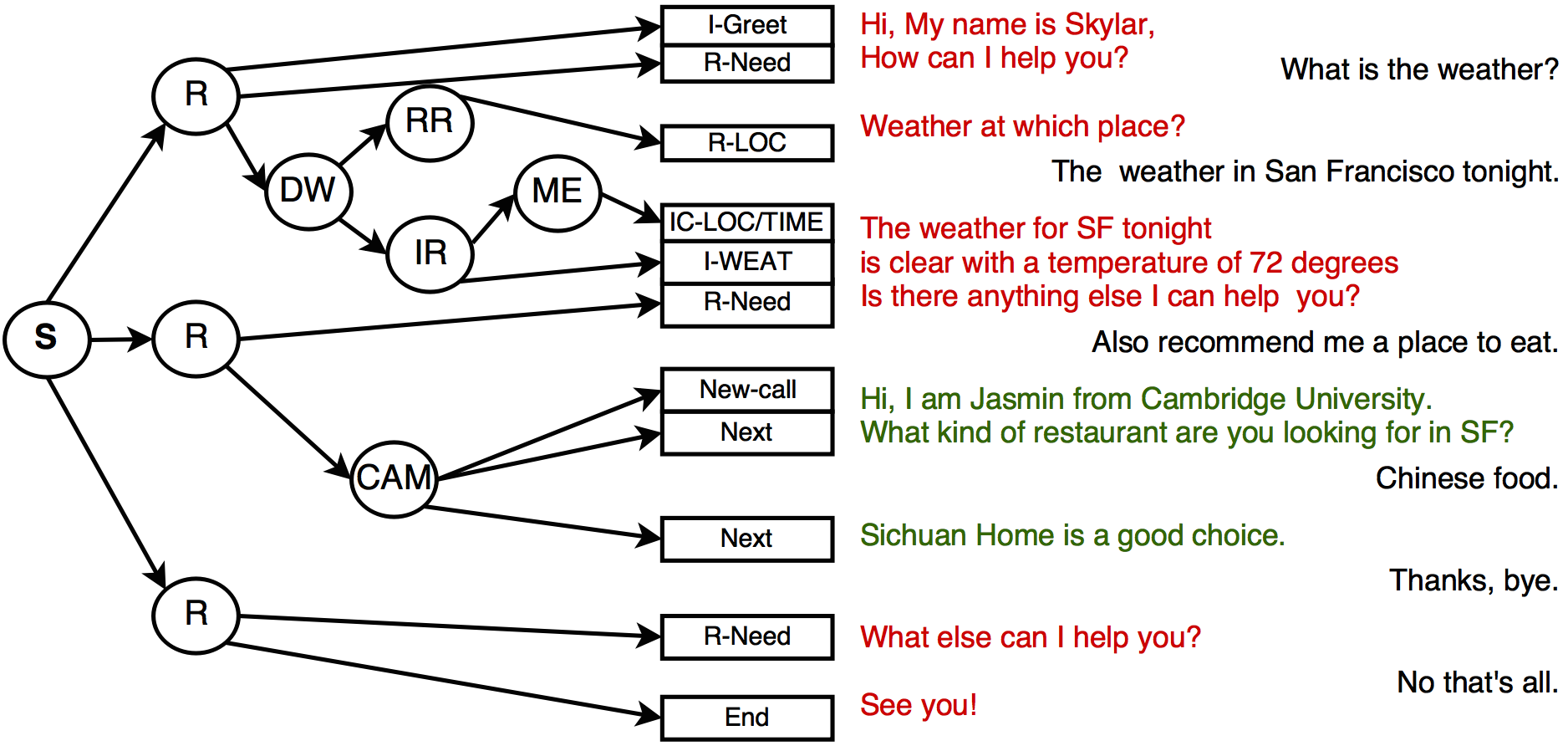}
    \caption{Example dialog with Skylar and the corresponding dialog tree. \textbf{S:} Start, \textbf{R:} Root, \textbf{DW:} Domain weather, \textbf{RR:} Request root, \textbf{IR:} Inform root, \textbf{ME:} Misunderstanding error, \textbf{CAM:} Cambridge Remote Agent. The \textcolor{red}{red} text is from Skylar and the \textcolor{green}{green} text is from the Cambridge Remote Agent.}
    \label{fig:parse}
\end{figure*}

\textbf{Hierarchical Reinforcement Learning:} The mathematical framework of HRL is the Markov Decision Process (MDP). An MDP is a tuple $(S, A, P, \gamma, R)$, where $S$ is a set of states; $A$ is a set of actions; $P$ defines the transition probability $P(s'|s,a)$; $R$ defines the expected immediate reward $R(s,a)$; and $\gamma \in [0, 1)$ is the discounting factor. Furthermore, an MDP $M$ can be decomposed into a set of subtasks, $O=\{O_1, O_2, ..O_k\}$, where $O_0$, by convention, is the root and solving $O_0$ solves $M$.  A subtask is a Semi-Markov Decision Process (SMDP), which is characterized by two additional variables compared to an MDP:
 \begin{enumerate}\setlength\itemsep{-\parsep}    
    \item $\beta_i(s)$ is the termination predicate of subtask $O_i$ that partitions $S$ into a set of active states, $S_i$ and a set of terminal states $T_i$. If $O_i$ enters a state in $T_i$, $O_i$ and its descendants exit immediately, i.e. $\beta_i(s)=1$ if $s\in T_i$, otherwise $\beta_i(s)=0$.
    \item $U_i$ is a nonempty set of actions that can be performed by $O_i$. The actions can be either primitive actions from $A$ or other subtasks, $O_j$, where $i\neq j$. We will refer to $U_i$ as the children of subtask $O_i$.
\end{enumerate}

Eventually a hierarchical policy for $M$ is $\pi$ and this is simply a set of policies for each subtask, i.e.  $\pi=\{\pi_0, \pi_1...\pi_n\}$. It is evident that a valid hierarchical policy forms a DAG, whose roots come from $O_0$ and whose terminal leaves are primitive actions belonging to $A$.

\textbf{ReinForest Dialog Policy:} ReinForest Dialog Policy forms a dialog task tree. Each node is a subtask belonging to one of three types: \textit{dialog agency}, \textit{dialog choice agency} and \textit{dialog agent}.
\begin{itemize}\setlength\itemsep{-\parsep}    
    \item Dialog Agency: a subtask $O_i \in O$ with a fixed policy that will execute its children from left to right. 
    \item Dialog Choice Agency: a subtask $O_i \in O$ with a learned policy that chooses the next executed child based on the context.
    \item Dialog Agent: a primitive action $a \in A$ that actually delivers the action to users. 
\end{itemize}

\subsection{Belief Update}
Belief update takes place when there is new input from the user. This component will first update the generic dialog states, such as incrementing the dialog turn count. Then it loops through all the concepts in the knowledge ontology and checks if the annotations in the new input match with any subscribed domain/intent/entities in each concept. If a match is found, the new values are stored in the concept's attribute map.

\subsection{Tree Transformation}
Tree transformation is key in ReinForest in order to support mixed-initiative and multi-domain dialogs. The transformation has two steps: candidate tree generation and candidate tree selection.

\textbf{Candidate tree generation:} scans through the updates made by \textit{belief update} and generates a list of candidate trees that can be pushed to the dialog stack (the list can be $\emptyset$). Usually a candidate is generated when the user explicitly requests the type of information that is handled by a different domain. 

\textbf{Candidate selection:} the selected candidate trees are appended under a dialog choice agency. The dialog choice agency selects one of the trees and pushes it to the dialog stack. 

\begin{table*}[t]
\centering
\caption{Characteristics of systems that handle non-goal driven utterances}
\label{table:non-task}
\resizebox{\textwidth}{!}{%
\begin{tabular}{|p{0.22\textwidth}||p{0.3\textwidth}|p{0.25\textwidth}|p{0.25\textwidth}|}
\hline
    & \textbf{Characteristics}   & \textbf{Main Technique}   &  \textbf{Representative Systems}                 \\ \hline
\textbf{(a) Pattern Matching}                   & high precision, fast response time, time consuming to make patterns  & Artificial Intelligence Markup Language~\cite{wallace2005aiml}    & Alice~\cite{wallace2009anatomy} \\ \hline
\textbf{(b) Example-based}           & high precision by threshold, slow response time if the data size is large &  vector space model~\cite{turney2010frequency}  & IRIS~\cite{banchs2012iris}  \\ \hline
\textbf{(c) Neural Chatbot}   & open domain, sometimes inconsistent and ungrammatical, require large corpora             & sequence-to-sequence learning~\cite{sutskever2014sequence} & CleverBot \cite{vinyals2015neural}   \\ \hline
%\textbf{(d) Knowledge Base Question Answering}          & Entity in KB       & High precision              & Structured data          & Semantic Parsing \cite{berant2013semantic}               \\ \hline
%\textbf{(e) Information Retrieval based Question Answering} & Sentence from web  & High recall                 & Search engine            & Information Retrieval               \\ \hline
\end{tabular}%
}
\end{table*}
\subsection{Error Handling}
There are two types of error handling: \textit{misunderstand error handling} and \textit{non-understand error handling}~\cite{bohus2003ravenclaw}. Misunderstand handling is used to conduct explicit or implicit confirms about concepts in the knowledge ontology. Specifically, the dialog engine will loop through all concepts in the \textit{user concept pool} and select concepts that are updated but not yet grounded for misunderstanding error handling. A misunderstanding subtask will then be pushed to the stack. The misunderstanding subtask will choose a built-in misunderstanding error handling strategy to confirm each concept. The current implementation supports two types of strategies: implicit and explicit confirm. \\
\indent On the other hand, non-understanding handling is activated when there is user input, but no update or tree-transformation is able to succeed. ReinForest implements a wide range of non-understanding handling strategies, ranging from the simple "can you repeat that?" to a response from an external chat-bot. 
\subsection{Execution Demonstration}
Figure~\ref{fig:parse} shows an example dialog with ReinForest. The dialog engine first pushes the \textit{root} on to the stack and asks what the user needs. After recognizing that the user is looking for weather information, it pushes the weather domain tree on to the stack. After acquiring all of the dependent concepts, Skylar informs the the user of the weather information. The user then decides to request restaurant information which is covered by the Cambridge Remote Agent so it is pushed to the stack. At this time, ReinForest transfers the control to Cambridge and simply calls \textit{next} to the remote agent in order to obtain the next response. When the remote agent terminates it's own session, Skylar takes back control and continues the conversation.

\section{Non-goal-driven Dialog Manager}
\label{sec:chatbot}
When a user's input cannot be handled by ReinForest, such as out-of-domain utterances (e.g., "you are smart"), specific questions (e.g., “who founded Microsoft?”, “how much is an iPhone?”), the non-understanding error handling policy triggers the non-goal driven DM to generate a system response. Goal-driven dialog systems focus on a set of predefined in-domain requests, non-goal driven dialog systems must handle open domain utterances which have in principle unlimited user intents. Most previous non-goal-driven approaches either used handcrafted pattern matching rules or example-based approaches that used a database manually designed by human experts. Recently, recurrent neural network-based data-driven approaches have been proposed that train on large movie transcription corpora (Table~\ref{table:non-task}). Most of the past approaches focus on building chat bots and do not provide a direct solution for DialPort, because we use the non-goal driven dialog manager together with goal driven dialog processing for both entertainment and error handling. Pattern-based approaches are expensive and time consuming (Table~\ref{table:non-task}-a). Moreover, existing publicly available patterns were designed to maintain a conversation with a chat bot for entertainment only. The Neural Network chat bots need a large amount of data to achieve good performance and their dialogs are biased towards the training data (e.g movie scripts). (Table~\ref{table:non-task}-c). Thus the learned dialog strategy is different from the one we want, which is used to recover non-understanding errors and to encourage users to speak to remote agents in other domains. 

The initial prototype of DialPort uses an example-based chat bot for the initial prototype, because the precision of the response can be controlled by a similarity threshold (Table~\ref{table:non-task}-b). We use the publicly available large knowledge base, Freebase\footnote{http://freebase.com} created by Google, to extend coverage. For example, if a user asks about a person, a location or the definition of a word, by using the Freebase ID extracted from the DBpedia spotlight and the Freebase property "common.topic.description", the system can find the requested information. Therefore, the non-understanding error handling policy queries the chat bot agent with the out-of-domain user input and the example-based chat bot calculates the similarity scores using sent2vec \cite{shen2014latent} (rather than a traditional vector space model). If the similarity score is over 80\%, the system response is selected from the chat bot agent. Otherwise ReinForest follows a deterministic error handling strategy which first asks users to "rephrase their request" and then provides more instructions if the error cannot be recovered.

\section{Integration Protocol}
\label{sec:protocol}
This section describes the integration protocol given to participants who are joining DialPort. This simple integration protocol concerns the text remote agent and the knowledge remote agent.

\textbf{Text Remote Agent:} a text remote agent is a \textit{dialog agency} defined in Section~\ref{sec:reinforest} and it only needs to implement two APIs:
\begin{itemize}
    \item $NewCall(id, s_0)$: The input parameters include the user $id$ and current dialog state $s_0$. The output is the first system response. The initial state $s_0$ enables the remote agent to skip redundant questions that were already asked in the previous conversation. DialPort calls this function to initialize a new session with the remote agent. Also, it is up to the remote research group how they use $s_0$, so the remote agent can operate totally independently. The exact format of $s_0$ can be customized if needed. 
    \item $Next(utt)$: The input is the users' utterance and the output is the system's response and an end-of-session flag. After $NewCall$, DialPort continues to call $Next$ to obtain the next system response until the end-of-session flag is $true$. Thus, the remote agent has complete autonomy during its session. We also note that the terminal flag is equivalent to the termination predicate $\beta_{i}(s)$ in the definition of an SMDP.
\end{itemize}
The purpose of DialPort is to collect and share real user data. So when a text remote agent finishes its session, it should be responsible for sending a dialog report along with the response to the last $Next$ call. The report should contain all the essential information about the conversation, such as the utterance at each turn. The final report format will be found on the DialPort website. Speech data that is collected will be made publicly available by the group who collected the data.

\textbf{Knowledge Remote Agent:} a knowledge agent is simply a function that outputs a list of matched entities, given a list of input constraints. Therefore, any common database format (e.g. SQL) or service API (e.g. Yelp API) can be a knowledge remote agent.

\section{Current State of DialPort}
\label{sec:current}
\begin{figure}[h]
\centering
    \includegraphics[width=6cm]{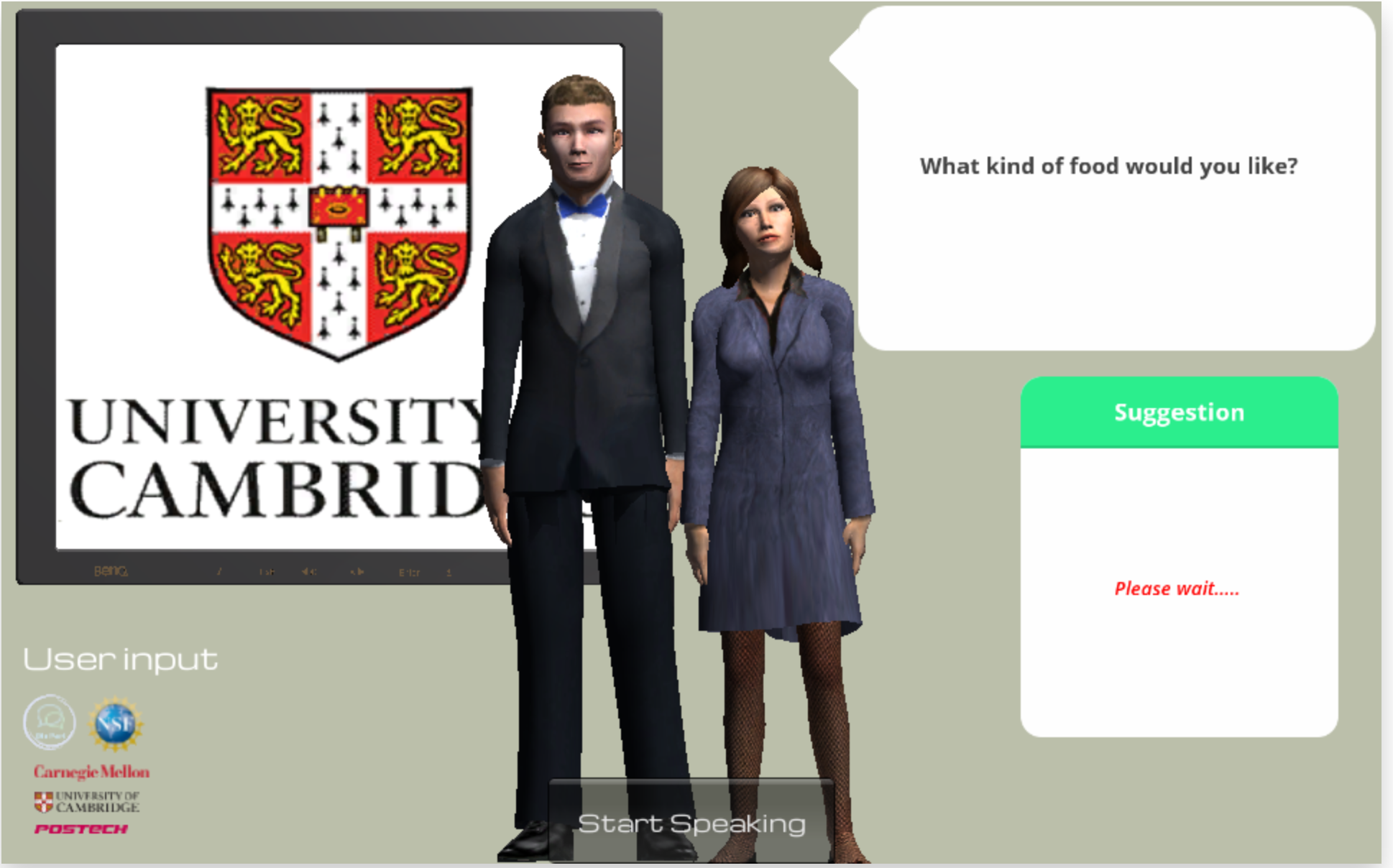}
    \caption{The present appearance of CMU-Skylar and Cambridge-Jasmin.}
    \label{fig:skylarAndJasmine}
\end{figure}

The first academic system that was connected to DialPort is from Cambridge University. In the near future, any academic system that can be connected will be welcome to join the portal. Figure~\ref{fig:skylarAndJasmine} shows the two current agents: a butler named Skylar from CMU and a librarian named Jasmin from Cambridge. Their appearance may change at a later time. DialPort will start to attract users to these systems as soon as it passes a series of stability tests and when several other remote agents, such as Yelp are added in order to broaden interest. CMU's Skylar will give information about the weather and restaurants other than in Cambridge and San Francisco. Its job is also to "sell" the other systems, getting the user to want to try them. When the domain changes, a new avatar appears and handles the conversation. When the dialog on the connected system’s topic is over, the user is handed back to Skylar. The transition is seamless from the users point of view. Jasmin from Cambridge gives information on hotels and restaurants in San Francisco. When the user is speaking with her, the logo in the background changes to the Cambridge log to indicate which system the user is speaking to (Figure~\ref{fig:skylarAndJasmine}).

\section{Conclusions}
\label{sec:conclude}
We propose a novel shared platform, DialPort, which can connect to many research systems enabling them to test new application ideas and gather real user data. In this paper, we have described the architecture of the user interface, DialPort, the virtual agents, and the (non)goal driven dialog managers and we have reported the current progress of the DialPort project. An important purpose of this paper is to encourage our colleagues to link their systems to DialPort so that we can help them to collect real user data.

\section{Acknowledgements}
This work was funded by NSF grant CNS-1512973. The opinions expressed in this paper do not necessarily reflect those of NSF. We would also like to thank Dialogue Systems Group at
Cambridge for their hard work in making the connection with DialPort.

\clearpage
\bibliographystyle{acl2016}
\bibliography{acl2016}
\end{document}